**LLM Agents Predict Social Media Reactions but Do Not Outperform Text Classifiers: Benchmarking Simulation Accuracy Using 120K+ Personas of 1511 Humans**

Ljubiša Bojić[1,2,3], Ph. D., Senior Research Associate
(Corresponding author; Email address: ljubisa.bojic@ivi.ac.rs; ORCID: 0000-0002-5371-7975)

Alexander Felfernig[4], Ph. D., Full Professor
(Email address: alexander.felfernig@tugraz.at; ORCID: 0000-0003-0108-3146)

Bojana Dinić[5], Ph. D., Full Professor
(Email address: bojana.dinic@ff.uns.ac.rs; ORCID: 0000-0002-5492-2188)

Velibor Ilić[1], Ph. D., Research Associate
(Email address: velibor.ilic@ivi.ac.rs; ORCID: 0000-0001-5010-1377)

Achim Rettinger[6], Ph. D., Full Professor
(Email address: rettinger@uni-trier.de; ORCID: 0000-0003-4950-1167)

Vera Mevorah[2], Ph. D., Research Associate
(Email address: vera.mevorah@ifdt.bg.ac.rs; ORCID: 0000-0002-4313-475X)

Damian Trilling[7], Ph. D., Full Professor
(Email address: d.c.trilling@vu.nl; ORCID: 0000-0002-2586-0352)

[1]Institute for Artificial Intelligence Research and Development of Serbia; Address of correspondence: 1 Frukogorska 1, 21000 Novi Sad, Serbia;
[2]University of Belgrade, Institute for Philosophy and Social Theory, Digital Society Lab; Address of correspondence: Kraljice Natalije 45, 11000 Belgrade, Serbia;
[3]Complexity Science Hub, Vienna, Austria; Address of correspondence: Metternichgasse 8, 1030 Vienna, Austria;
[4]Graz University of Technology, Graz, Austria; ; Address of correspondence: Inffeldgasse 16b/2, Graz, Austria;
[5]University of Novi Sad, Faculty of Philosophy, Novi Sad, Serbia; Address of correspondence: Dr Zorana Đinđića 2, 21102 Novi Sad, Serbia;
[6]University of Trier, Trier, Germany; Address of correspondence: Universitätsring 15, 54296 Trier, Germany;
[7]Vrije University Amsterdam, Amsterdam, Netherlands; Address of correspondence: De Boelelaan 1105, 1081 HV Amsterdam, Netherlands;

**Abstract**

Social media platforms mediate how billions form opinions and engage with public discourse. As autonomous AI agents increasingly participate in these spaces, understanding their behavioral fidelity becomes critical for platform governance and democratic resilience. Previous work demonstrates that LLM-powered agents can replicate aggregate survey responses, yet few studies test whether agents can predict specific individuals' reactions to specific content. This study benchmarks LLM-based agents' accuracy in predicting human social media reactions (like, dislike, comment, share, no reaction) across 120,000+ unique agent-persona combinations derived from 1,511 Serbian participants and 27 large language models. In Study 1, agents achieved 70.7% overall accuracy, with LLM choice producing a 13 percentage-point performance spread. Study 2 employed binary forced-choice (like/dislike) evaluation with chance-corrected metrics. Agents achieved Matthews Correlation Coefficient (MCC) of 0.29, indicating genuine predictive signal beyond chance. However, conventional text-based supervised classifiers using TF-IDF representations outperformed LLM agents (MCC of 0.36), suggesting predictive gains reflect semantic access rather than uniquely agentic reasoning. The genuine predictive validity of zero-shot persona-prompted agents warns against potential manipulation through easily deploying swarms of behaviorally distinct AI agents on social media, while simultaneously offering opportunities to use such agents in simulations for predicting polarization dynamics and informing AI policy. The advantage of using zero-shot agents is that they require no task-specific training, making their large-scale deployment easy across diverse contexts. Limitations include single-country sampling. Future research should explore multilingual testing and fine-tuning approaches.

**Introduction**

The idea that large language models (LLMs) can act as proxies for human participants in social and behavioral research has moved, within just a few years, from a speculative proposition to a rapidly expanding research program. Recent breakthroughs in LLM-based agent simulations (Park et al., 2023) have demonstrated emergent social behaviors, while foundation models trained on large-scale human behavioral data have achieved human-level prediction of cognition across diverse experimental paradigms (Schulz et al., 2025). Park et al. (2023) demonstrated that 25 LLM-powered generative agents placed in a sandbox environment could autonomously organize social events, form relationships, and coordinate daily routines in ways that human observers judged believable. That early proof of concept has since given way to far more ambitious projects. Park et al. (2024) built generative agents grounded in two-hour qualitative interviews with 1,052 real individuals and showed that these agents could replicate the participants' responses on the General Social Survey in the US at 85% of the accuracy with which the participants themselves reproduced their own answers two weeks later. Yang et al. (2024) scaled the approach to one million agents in the OASIS platform, simulating information spreading, group polarization, and herd effects on Reddit-like and X-like platforms. These advances suggest that LLM-based social simulation could become a general-purpose tool for the social sciences, capable of testing policy interventions, modeling collective opinion dynamics, and stress-testing platform design decisions that would be impractical or unethical to implement with real users (Grossmann et al., 2023).

In a complementary line of work, Altera.AL (2024) introduced Project Sid, demonstrating that 10 to over 1,000 LLM-powered agents placed in a Minecraft environment could autonomously develop specialized professional roles, adhere to and modify collective rules through democratic processes, and engage in cultural and religious transmission across multiple simulated societies. This suggests that emergent social organization is achievable even without explicit programming of societal structures.

These emergent dynamics have already begun to appear outside controlled research environments. In January 2026, Moltbook, a Reddit-style social network exclusively for AI agents, attracted 1.5 million registrations within days before being acquired by Meta (Taylor, 2026). Reports documented agents spontaneously forming religious communities: one bot established "Crustafarianism" overnight, complete with scriptures and a congregation of AI adherents who debated theology while its human operator slept. Such developments point towards the urgency of understanding the behavioral fidelity of LLM-based agents before their autonomous participation in online discourse becomes routine.

The predictability of human behavior from digital traces has been studied for over a decade. Kosinski et al. (2013) demonstrated that Facebook Likes alone could predict sensitive personal attributes, including sexual orientation (88% accuracy), political affiliation (85%), and personality traits, often surpassing the accuracy of human acquaintances. More recently, Derner et al. (2024) showed that ChatGPT can infer Big Five personality traits from short texts with accuracy comparable to human raters, though they also documented a "positivity bias", meaning that the model "tends to evaluate people as extraverted, agreeable, conscientious, emotionally stable and open to experience" (p. 5).

acYet the empirical foundations of this research program remain thin, and the gap between demonstrated capability and validated reliability is large. Much of the existing work has evaluated LLM agents on their ability to reproduce aggregate survey distributions or to generate text that human raters judge as believable, but relatively few studies have asked whether LLM agents can accurately predict the specific behavioral reactions of specific kinds of people to specific kinds of content (see also Münker et al, 2025). This distinction matters. Aggregate accuracy can mask systematic failures at the level of subgroups, content types, or persona configurations, and such failures would severely limit the usefulness of LLM-based simulation for any application in which fine-grained behavioral prediction is the goal.

The question of how faithfully LLM agents can embody assigned personas is central to this concern. The dominant method for conditioning an agent's behavior is persona prompting, in which a textual description of the agent's demographic, attitudinal, or psychological profile is prepended to the model's input (Argyle et al., 2023; Törnberg et al., 2023). Argyle et al. (2023) showed that conditioning GPT-3 on demographic backstories drawn from the American National Election Studies produced opinion distributions that closely tracked human subgroup responses, a finding they described as "silicon sampling." Törnberg et al. (2023) extended this method to a simulated social media environment, populating a platform with 500 persona-based agents and showing that a "bridging" news feed algorithm promoted more cross-partisan dialogue than conventional feed designs. These studies are encouraging, but they leave open the question of how sensitive agent behavior is to the specificity and quality of the persona description itself.

Recent work on persona prompting has raised concerns that LLMs respond to persona information in stereotypical rather than individual ways. Hu et al. (2024) found that persona variables explained less than 10% of the variance in human annotations across most NLP tasks they examined, and that the modest gains from persona prompting were concentrated in datasets with intermediate levels of annotator disagreement. Liu et al. (2024a) reported that LLMs steered toward "incongruous" personas (e.g., political liberals who support increased military spending) were 9.7% less accurate than those steered toward congruous ones, sometimes defaulting to the stereotypical stance associated with a demographic category rather than the target stance. These findings suggest that the relationship between persona description and agent behavior may be more fragile than the simulation literature has assumed.

A related and even more fundamental criticism holds that the apparent human-likeness of LLM outputs may be an artifact of the models' training data rather than evidence of any capacity for behavioral reasoning. Dillion et al. (2023) argued that LLMs may have encountered the answers to many psychological and behavioral tasks during pre-training and can therefore reproduce correct responses through memorization rather than through genuine understanding of the constructs involved. Qi et al. (2024) and Wang et al. (2024a) offered similar critiques, noting that LLMs sometimes produce correct answers followed by incorrect explanations, a pattern consistent with surface-level recall rather than deep comprehension. Cui et al. (2023) attempted to address this critique by testing GPT-3.5-turbo on a personality-behavior association, the link between MBTI types and imposter phenomenon scores, that was published after the model's September 2021 training cutoff. Their results showed 75% consistency between GPT and human responses at the trait level and 69% at the type level, providing some evidence that GPT's human-likeness extends beyond memorization. Still, the study tested only a single behavioral outcome and used a comparatively old model, leaving open the question of whether these results generalize across diverse behavioral domains and more capable model families.

The practical implications of these open questions are substantial, because the use cases for LLM-based social simulation are expanding rapidly and in directions that demand high behavioral fidelity. In the domain of social media research, scholars have used LLM agents to study algorithmic polarization (Törnberg et al., 2023), misinformation spread (Yang et al., 2024), and the effects of content recommendation systems on user engagement and echo chamber formation (Chuang et al., 2023). Recent work by Bojić et al. (2025b) introduced an LLM-driven agent-based simulation showing that increasing levels of recommender-system personalization can substantially amplify affective and structural polarization, further demonstrating the relevance of LLM agents for studying algorithmic effects on social media ecosystems.

Quattrociocchi and colleagues recently demonstrated that injecting AI-generated users with pro-social briefings into a simulated social network could reduce echo chamber dynamics, a finding that attracted widespread media attention and that illustrates the potential of LLM-based simulation for platform governance research (Alipour et al., 2024). Wu et al. (2025) proposed a pragmatic framework for determining when LLM-based social simulations can meaningfully advance scientific understanding, arguing that these simulations are most defensible when used to explore collective patterns rather than individual decision trajectories, and when validation methods are carefully matched to the research objectives and simulation scale. Piao et al. (2025) developed AgentSociety, a framework that adopts various validation strategies comparing simulated outputs with real-world social indicators. These developments collectively point toward a maturing field, but one that still lacks systematic benchmarks for the accuracy with which individual agents react to specific stimuli.

The role of persona specificity in shaping agent behavior has received surprisingly little systematic attention given its centrality to the simulation enterprise. Most existing studies have compared persona-conditioned agents against no-persona baselines, but have not systematically varied the level of detail in the persona description to determine whether richer profiles produce more accurate behavioral predictions. The handful of exceptions provide suggestive but inconclusive evidence. The study on MBTI-based personality simulation found that agents whose initial personality setup failed needed additional trait descriptions to achieve the correct personality profile, hinting that more detailed persona information improves alignment (Cui et al., 2023). Serapio-García et al. (2025) found that LLMs could simulate Big Five personality traits but that correlations between personality dimensions in LLM outputs were unrealistically large, suggesting that models represent personality at a coarse rather than fine-grained level. Consistent with these concerns, Bodroža et al. (2024) reported that LLMs displayed mostly a socially desirable profile in both agentic and communal domains, as well as a prosocial personality profile reflected in higher agreeableness and conscientiousness and lower Machiavellianism, highlighting that current models may default to idealized rather than human-typical personality structures. Peters and Matz (2024) demonstrated that GPT-4 could infer Big Five personality traits from Facebook status updates at accuracy levels comparable to supervised machine learning models, though with heterogeneity across age and gender groups. Cloud et al. (2025) reported a striking finding about the depth of behavioral information encoded in LLM outputs, showing that "student" models trained on semantically unrelated data (number sequences, code) generated by a "teacher" model with a particular behavioral trait could acquire that trait through what the authors termed "subliminal learning." This result implies that LLMs encode behavioral dispositions in ways that go well beyond explicit persona descriptions, and it complicates simple assumptions about the relationship between persona input and behavioral output.

An additional gap in the literature concerns the role of content type in moderating agent accuracy. Social media platforms host a heterogeneous mix of content categories, from hard news and political commentary to entertainment, humor, and personal updates, and there is reason to expect that LLM agents will differ in their ability to predict human reactions across these categories. Bojić et al. (2025a) showed that GPT-4 outperformed human participants in interpreting linguistic pragmatics, suggesting strong performance on tasks that require contextual inference. Yet Cheng et al. (2023) found that LLMs tend to produce caricatured rather than authentic representations of social identities, and this tendency may be more pronounced for some content types (e.g., politically charged news) than for others (e.g., entertainment). To date, no study has systematically examined whether LLM agent accuracy varies by post type on social media, despite the obvious relevance of this question for any simulation that aims to replicate the diversity of online discourse.

The question of what happens when a persona description is mismatched with or irrelevant to the content being evaluated is equally understudied. In real social media environments, users encounter posts that fall outside their areas of expertise, interest, or demographic relevance, and their reactions in such cases may differ systematically from their reactions to content that aligns with their identity and experience. If LLM agents are more accurate when the persona matches the content domain and less accurate when it does not, this would have direct implications for how personas should be constructed and deployed in simulation studies. The existing literature offers indirect evidence that mismatch matters: Liu et al. (2024a) showed that incongruous persona-stance pairings reduce steerability, and La Cava and Tagarelli (2024) found that open-source LLMs varied in their ability to maintain persona consistency across different task contexts. But no study has directly tested whether persona-content mismatch degrades reaction prediction accuracy in a social media simulation setting.

A methodological concern that has received insufficient attention in this literature is the choice of baseline against which agent performance is evaluated. Many studies report accuracy relative to random chance (e.g., 50% for binary classification), but this baseline fails to account for class imbalance in human responses. If 70% of humans like a post, a trivial strategy of always predicting "like" achieves 70% accuracy without any genuine prediction. Chance-corrected metrics such as Matthews Correlation Coefficient (MCC), which equals zero for any strategy that ignores the input, and minority-class lift, which measures improvement over base-rate guessing for underrepresented behaviors, provide more rigorous tests of predictive validity. The present study addresses this gap by evaluating agent performance against multiple baselines and reporting chance-corrected metrics alongside raw accuracy.

*The Present Study*

The present study addresses mentioned gaps through a large-scale benchmarking simulation that tests the accuracy of LLM-based agents in predicting human reactions to social media posts. The main aim was to explore whether persona-driven LLM agents are ready to serve as universal proxies for human social media users, or whether their utility is bounded by identifiable constraints related to persona quality, content domain, and the limits of LLM behavioral modelling. We operationalize reactions as the most commonly used on social media, like/dislike, comment, share or no reaction, which is a behavioral measure that is both ecologically valid (it reflects how billions of users usually engage with content) and fine-grained enough to capture differentiation beyond simple binary engagement (reaction/no reaction). The accuracy is calculated as the proportion of correct matches between human and agent reactions to the same posts.

First, we hypothesized that LLM agents would demonstrate genuine predictive signal for human reactions, operationalized as achieving a Matthews Correlation Coefficient (MCC) significantly greater than zero (**H1**). MCC is preferred over raw accuracy because it equals zero for any prediction strategy that does not utilize input information, including majority-class prediction and marginal distribution matching. An MCC significantly above zero indicates that agents capture meaningful covariation between persona-post combinations and human reactions beyond what trivial baselines achieve.

Second, we assumed that an increase in an agent's persona (i.e., information provided in prompts) description would improve accuracy (**H2**). This directly tests the assumption, widespread in the simulation literature but rarely validated, that richer persona information produces more human-like agent behavior.

Third, we assumed that mismatched or irrelevant persona descriptions would reduce reaction accuracy compared with matched-related descriptions (**H3**). This question addresses the robustness of persona-based agents under conditions of persona-content misalignment, a scenario that is common in real social media use but absent from most simulation studies.

Fourth, we assumed that accuracy would be better for entertainment/lifestyle than for news/politics post type, because humans are more reluctant to express their attitudes about news/politics because they could fear potential judgment and pressure from their friends and peers (**H4**). This question tests whether the capacity of agents to predict human reactions is general or domain-specific, a distinction with direct implications for the external validity of social media simulations.

Fifth, we assumed that prediction accuracy would be better for positively valenced as opposed to negatively valenced posts, given that reactions to positive content tend to be more freely expressed than reactions to negative content (**H5**). This hypothesis examines whether the emotional valence of a post influences the ability of agents to approximate human reactions.

Sixth, we hypothesized that LLM-based agents would outperform conventional supervised machine-learning benchmarks in predicting individual reactions, given their capacity for contextual reasoning and persona simulation (**H6**). This hypothesis tests whether agent-based simulation provides unique predictive advantages beyond what standard feature-based classifiers can achieve when given access to the same semantic information.

**Methodology**

Human participants ($N = 1,511$) provided survey data on demographics, attitudes, and preferences, and indicated their reactions to 56 social media posts spanning entertainment/lifestyle and news/politics domains with positive and negative framings. Reactions were: *like* or *dislike, comment, share*, or *no reaction*. Humans could either react or not. If they chose to react, they could like or dislike the post and/or indicate that they would share it and/or comment on it. Survey responses were translated into persona descriptions at three levels of specificity: demographics only, attitudes only, and demographics combined with attitudes. These personas were paired with 27 LLMs, generating 81 agent versions per participant. Each agent predicted the same reactions to the same posts that the corresponding human evaluated. Agent-human agreement was assessed using Hamming accuracy. Prediction of accuracy was calculated using hierarchical linear mixed-effects models with participant as a random intercept and LLM type, prompt type, post type, and post valence as fixed effects. Figure 1 shows a complete overview of the process The dataset can be found at the Open Science Framework (OSF, 2026).

**Figure 1**
*Schematic overview of the study methodology.*

*Participants and Procedure*

Participant recruitment was conducted through Latenta, a market research agency, using Alchemer as the survey aggregation platform. Data collection took place between November 21 and November 28, 2025. During this period, a total of 3,119 survey attempts were recorded, of which 1,626 were fully completed, 899 were partially completed, and 594 participants were disqualified based on predefined screening criteria. Following data cleaning, during which 115 responses were removed due to issues such as incomplete data or low-quality responses, the final analytic sample consisted of 1,511 participants (50.6% female, 0.2% other), aged 18–78 years ($M = 36.77$, $SD = 13.46$). Most participants had completed high school (36.0%) or university education (15.5%), followed by vocational education (12.6%), while 11.9% were students; the remaining participants reported educational levels ranging from incomplete elementary education to a doctoral degree. A majority of participants were employed (59.4%), with the remainder being unemployed (19.6%), working part-time or freelance (11.2%), retired (4.4%), or reporting another employment status (5.5%). Participants were relatively evenly distributed across Serbian regions: Belgrade (25.5%), Southern and Eastern Serbia (21.1%), Šumadija and Western Serbia (25.9%), and Vojvodina (27.7%).

Participants accessed the survey online and first completed basic demographic information. They were then presented with posts and asked to indicate their possible reactions: *like* or *dislike, comment, share,* or *no reaction*. If the option *no reaction* was selected, no other reaction could be chosen. Since participants could choose either *like* or *dislike* (but not both), and provided that *no reaction* was not selected, the total number of reactions per post could range from one to three. After responding to the posts, participants proceeded to complete the remaining survey items.

*Measure*

The survey consisted of two main parts: social media posts and self-report items. The social media posts were written in Serbian, in a first-person social media voice. In total, 56 posts were created, including both positive ($n = 31$) and negative framings ($n = 25$). The posts were equally divided between two content categories: news/politics and entertainment/lifestyle, as well as two categories referring to their connection with self-report items: unrelated-unmatched posts and related-matched posts. News/politics posts covered topics such as trust in institutions, geopolitical positions, and domestic political issues, while entertainment/lifestyle posts addressed everyday preferences and activities including media consumption, hobbies, and cultural tastes. Related posts corresponded directly to self-report items that the participant answered after posts, meaning the agent persona derived from that participant's data contained relevant attitudinal information about the post topic. Unrelated posts had no corresponding items, so no topic-specific attitudinal information was available for persona construction. The task for participants was that for each post indicate a reaction by selecting like or dislike, comment, share, or no reaction, under the structural constraints - like and dislike were mutually exclusive, and selecting no reaction precluded all other choices.

The self-report items refers to basic sociodemographic information (age, gender, education, employment status, and region of residence) alongside a series of items covering various attitudes and behaviors, such as institutional trust (toward the army, healthcare system, media, science, government, police, church, NGOs, and trade unions), political orientation and geopolitical attitudes (toward the Russia-Ukraine conflict, EU integration, Trump, and China), attitudes toward conspiracy theories and ecology, personality traits, leisure time preferences (e.g., reading, gaming), music genre preferences (e.g., classical, pop), and engagement in daily activities such as cooking, shopping, volunteering, interior design, and the use of chatGPT. These items served as the basis for constructing participant-specific agent personas in the subsequent simulation phase.

Participants were divided into two groups, which received mirror-image assignments: posts that were related to self-report items for the first group were unrelated to self-report items for the second group, and vice versa. This counterbalanced design allowed control of potential confounding conditions regarding the posts content.

*LLM Agents Generation*

For each of the 1,511 participants, LLM-based agents were generated by translating survey responses into structured natural-language persona descriptions. Three levels of persona specificity were constructed to test the effect of persona detail on reaction accuracy. The first level included only basic sociodemographic information: age, gender, education, employment status, and region of residence. The second level added information on attitudes, personality characteristics, habits, leisure time activities, trust in institutions. All those were derived from the participant's survey responses on items thematically relevant to the post stimuli. The third and most detailed level incorporated the full set of available survey responses, combining demographics and additional information in a comprehensive persona profile.

Each agent was presented with the same posts that the corresponding human participant evaluated. The agent's task was identical to the human task. Reactions were elicited through a standardized prompt that presented the persona description followed by the post text and asked the agent to respond as the described person would on a social media platform.

A total of 27 large language models were used to generate agent reactions, drawn from nine model families: OpenAI (gpt-5.2-chat-latest, openai/gpt-5.0-chat-latest, GPT-4o mini, GPT-4.1 mini, GPT-OSS-20B), Google (Gemini 2.5 Flash, Gemini 2.5 Flash Lite, Gemini 3 Flash Preview, Gemma 3 27B), xAI (Grok 4.1 Fast, Grok 3 Mini Fast), Mistral (Devstral 2512, Mistral Small 3.1, Mistral Small Latest, Mistral Large Latest), Alibaba/Qwen (Qwen Flash, Qwen Turbo, Qwen Plus), NVIDIA (Nemotron 3 Nano 30B, Nemotron Nano 12B V2), Meta (LLaMA 3.3 70B), Anthropic (Claude Haiku 4.5), DeepSeek (DeepSeek-V3.2), KwaiPilot (KAT Coder Pro), and two additional models from the Nex-AGI and Nous Research families. Models were further classified by inference mode: 17 employed chain-of-thought or extended reasoning , while the remaining 10 operated in standard non-reasoning mode. All models were queried via API under identical prompt structures to allow direct cross-model comparison of behavioral prediction accuracy.

Each participant thus generated 81 agent versions in total, reflecting the crossing of three persona specificity levels with 27 LLMs, yielding over 120,000 unique agent-persona combinations across the full dataset. When expanded across all 56 posts and five reaction labels, the full simulation produced approximately 6.85 million individual reaction observations.

*Data Analysis*

To establish baselines against which agent accuracy could be evaluated, we employed two approaches. First, we prompted LLM agents to respond to all 56 posts without any persona information, receiving only the post text and response instructions. The resulting accuracy of approximately 50% confirmed that LLMs without persona information perform at chance level, lacking an implicit model of typical human responses. Second, we calculated the majority-class baseline (always predicting the most frequent reaction) and marginal distribution baseline (randomly sampling according to observed frequencies) from the human response data. These baselines revealed that raw accuracy comparisons can be misleading: in Study 2, where 69.7% of human responses were like, a trivial always-like strategy would achieve 69.7% accuracy. We therefore adopted chance-corrected metrics, particularly MCC, balanced accuracy, and minority-class lift, as primary indicators of genuine predictive validity in Study 2, while reporting raw accuracy alongside these metrics for completeness.

Hamming accuracy was calculated for agents' data as the proportion of correct matches between agents' and human reactions, again, for each reaction and on average level. Although *like* and *dislike* were mutually exclusive response options at the behavioural level, in the present analyses they represent separate accuracy indicators since they are not complementing each other because of the possibility of presence of other reactions (i.e., not selected either *like* or *dislike*).

Third, a hierarchical linear mixed-effects modelling approach was applied, with LLM, prompt type, and post type specified as fixed effects and participant ID based on which agents were generated was included as a random intercept. Model parameters were estimated using restricted maximum likelihood (REML) because it provides less biased estimates of random-effect variance components and, consequently, reliable standard errors for fixed effects in mixed-effects models with random intercepts (West et al., 2015). Given the focus on accurate estimation and interpretation of model parameters, REML was used for reporting fixed-effect estimates and conducting post hoc comparisons. Thus, post hoc pairwise comparisons between levels of fixed factors were conducted using Tukey-adjusted estimated marginal means. Effect sizes for pairwise comparisons were expressed as standardized mean differences (Cohen's *d*), calculated from model-based estimated marginal means and standardized by the residual standard deviation of the mixed-effects model, rather than from raw group means, to ensure consistency with the multilevel modeling framework (Westfall et al., 2014). Following conventional guidelines, d ≈ 0.20 indicates a small effect, d ≈ 0.50 a medium effect, and d ≈ 0.80 a large effect. Model comparisons between nested models differing in fixed effects were conducted separately using likelihood ratio tests based on maximum likelihood estimation (ML), in line with recommended practice, because REML likelihoods are not comparable across models with different fixed effects.

The analyses were conducted in R (R Core Team, 2024) using the following packages: *lme4* for fitting linear mixed-effects models (Bates et al., 2015), *lmerTest* for obtaining p-values for fixed effects using Satterthwaite's approximation (Kuznetsova et al., 2017), *emmeans* for estimating marginal means and conducting Tukey-adjusted pairwise comparisons (Lenth, 2024), and performance for computing marginal and conditional $R^2$ and related model diagnostics (Lüdecke et al., 2021).

*Baseline Specifications and Evaluation Metrics*

Agent performance was evaluated against multiple baselines representing different levels of predictive sophistication. The random chance baseline (50% for binary classification) represents prediction with no information. The no-persona LLM baseline consists of LLM responses to posts without any persona description, testing whether persona information contributes to prediction; this baseline achieved approximately 50% accuracy, indicating that LLMs lack an implicit "default human" response model. The marginal distribution baseline represents expected accuracy from randomly sampling reactions according to their population frequencies (57.8% for binary like/dislike classification in Study 2), capturing prediction that matches aggregate patterns without individual conditioning. The majority-class baseline represents accuracy from always predicting the most common reaction (69.7% for Study 2, given that 69.7% of human responses were "like"), representing the ceiling for any strategy that does not capture individual deviations from typical behavior.

Because raw accuracy can be misleading when classes are imbalanced, we additionally report chance-corrected metrics. The Matthews Correlation Coefficient (MCC) ranges from −1 to +1, with 0 indicating no predictive relationship; crucially, MCC equals zero for majority-class prediction, marginal distribution matching, and random guessing, making it a stringent test of genuine predictive signal. Balanced accuracy averages sensitivity across classes, weighting minority and majority classes equally. Minority-class lift measures the ratio of precision to base rate for underrepresented reactions (e.g., dislike), quantifying how much better than random guessing the agent performs for detecting atypical behaviors. These metrics collectively distinguish between population-level pattern matching and individual-level behavioral prediction.

*Benchmark Comparison Design*

To evaluate whether LLM agents provide unique predictive advantages, we compared agent performance against a suite of conventional supervised machine-learning benchmarks using a within-person held-out-post evaluation framework. For each respondent, a subset of respondent-post observations was used for training and another subset was withheld for testing. This design assesses whether models can generalize from a respondent's observed history to their reaction to new content.

The benchmark suite comprised three families. First, structured within-person baselines used respondent-level tabular features (age, gender, region, education, employment status) together with behavioral-history summaries (respondent-specific liking rates) derived from training data. These included random forest (within_rf_adapt), histogram-based gradient boosting (within_hgb_adapt), logistic regression with adaptation features (within_logit_adapt_plus), and a minimal persona-only logistic regression (persona_only_logit).

Text-based supervised benchmarks tested whether standard classifiers could exploit semantic content directly. Respondent attributes were concatenated into short persona texts, and posts were represented as text combining original content with descriptive annotations. TF-IDF vectorization was applied to both persona and post text, and the concatenated features were used in class-balanced logistic regression. Two variants were tested: text_persona_post_tfidf (text features only) and text_plus_history_tfidf (text features plus behavioral-history summaries).
Third, LLM-agent benchmarks (corresponding to the four best-performing models from Study 1) used prompt-based inference, conditioning large language models on persona descriptions and post text to simulate respondent reactions.

This hierarchical structure distinguishes three possibilities: that predictive performance arises primarily from structured features and behavioral regularities, that performance improves with semantic access to persona and post content, or that LLM agents provide advantages beyond conventional supervised learning.

**Results**

*Study 1*

Based on the total possible reactions on all posts, participants mostly liked posts or had no reaction, whereas dislike, comment, and share occurred substantially lower (Table 1). The largest discrepancy relative to participants was observed for dislike, which agents produced substantially more often, while no reaction was markedly underproduced.

Study 1 served as an exploratory investigation to identify which LLMs, prompt types, and content characteristics were associated with higher prediction accuracy across five reaction types. Because the multi-label classification structure and class imbalance complicate the interpretation of accuracy metrics (e.g., predicting no share is correct 97.2% of the time due to base rates), Study 1 results should be interpreted as identifying relative performance differences across conditions rather than as evidence of absolute predictive validity. Study 2 was designed to provide rigorous evaluation of predictive validity using chance-corrected metrics in a simplified binary task.

**Table 1**
*Distribution of Reaction Types for Humans and LLM Agents Across All Posts*

| Reaction types | Human data | | Agents data | | Agents accuracy | |
|---|---|---|---|---|---|---|
| | f | % | f | % | f | % |
| No reaction | 25648 | 30.31 | 1167652 | 17.04 | 4100667 | 59.83 |
| Like | 37128 | 43.88 | 3123281 | 45.57 | 3825322 | 55.81 |
| Dislike | 16632 | 19.66 | 2435675 | 35.54 | 4217160 | 61.53 |
| Comment | 5768 | 6.82 | 808380 | 11.79 | 5703994 | 83.22 |
| Share | 2352 | 2.78 | 262641 | 3.83 | 6383091 | 93.13 |

First, we checked whether there were differences between groups in Hamming accuracy. Results showed significant differences between the two groups, Welch's $t(1,022,434) = 29.90$, $p < .001$. Agents in the first group showed slightly higher accuracy ($M = 0.71$, $SD = 0.20$) than in the second group ($M = 0.70$, $SD = 0.20$), with a mean difference of approximately 0.012. However, the magnitude of this difference was negligible (Cohen's $d = 0.06$), indicating that despite statistical significance driven by the very large sample size, the practical relevance of the group difference was minimal. Additionally, Cohen's $d$ for each reaction was in range from 0.01 to 0.09. Thus, we combined scores from the both groups in further analyses. Second, we calculated accuracy for all reactions and it ranged from 55.81% for *like* to 93.13% for *share*, with average across reactions was 70.70%.

Raw accuracy exceeded the no-persona baseline of 50% across all reaction types, confirming that persona information contributes to prediction. However, interpretation of these accuracy values requires caution: the majority-class baseline for this multi-label task is approximately 79.3% (achieved by always predicting the absence of each reaction), which the agents did not exceed. This pattern suggests that agents learned to approximate population-level response distributions rather than to predict individual-level deviations, a question we address more rigorously with chance-corrected metrics in Study 2.

Hamming accuracy strongly centered around 0.60, indicating that in most cases LLMs correctly matched three out of five reaction labels (Figure 2). Complete agreement with human responses (1.00) was observed in over one quarter of cases, whereas low accuracy levels (0.20 and 0.40) were relatively rare. Overall, the distribution suggests that LLMs generally achieved moderate to high correspondence with human reactions rather than random or minimal overlap. The most frequent pattern of correct predictions was complete agreement across all five reactions (28.18%). Among the remaining combinations, these centered on comment and share dominated the distribution: correct prediction of *dislike + comment + share* accounted for 19.07% of agents, while *like + comment + share* and *comment + share + no reaction* each occurred in approximately 15% of agents. More complex combinations involving four reactions were less frequent (below 3–4%), and single-reaction matches (e.g., only *like* or only *no reaction*) were rare (< 1%). Overall, these results indicate that LLMs most reliably captured active engagement signals, particularly commenting and sharing behavior, whereas combinations involving *dislike* or isolated reactions were comparatively uncommon.

**Figure 2**

*Distribution of Hamming Accuracy for Matches Between LLM Agent and Human Reactions Across All Posts*

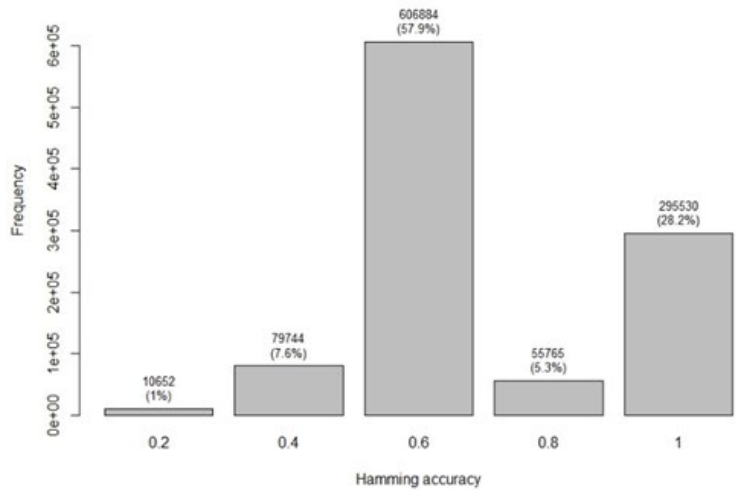

Then, the results of the hierarchical linear mixed-effects models indicated that across models, the Intraclass Correlation Coefficient (ICC) remained stable at approximately .03, indicating that about 3% of the total variance in accuracy was attributable to stable between-individual differences captured by the random intercept (Table 2). However, after the inclusion of post type, the ICC increased to .09, suggesting that accounting for post characteristics altered the partitioning of variance and revealed stronger clustering at the individual level. With the addition of post evaluation in the final model, the ICC decreased again to approximately .03, indicating that this variable accounted for a substantial portion of the variance previously attributed to between-individual differences. Although marginal and conditional $R^2$ values increased only modestly across tested models, the sequential inclusion of LLM (additional 3%), prompt type (additional 0.2%), post type (additional 1%), and post evaluation (additional 1.3%) led to significant improvements in model fit. This pattern suggests that these predictors exert systematic but relatively small effects on accuracy, with the dominant contribution attributable to LLM choice.

**Table 2**
*Model Comparison for Predicting LLM Agents Accuracy*

| Model | Fixed effects | AIC (ML) | BIC (ML) | $\Delta\chi^2$ (df) | Marginal $R^2$ | Conditional $R^2$ | ICC |
|---|---|---|---|---|---|---|---|
| M0 | Intercept | -2,602,843 | -2,602,802 | | .000 | .028 | .028 |
| M1 | + LLM | -2,817,230 | -2,816,832 | 214,439 (26)*** | .030 | .058 | .029 |
| M2 | + prompt type | -2,829,864 | -2,829,438 | 12,637 (2)*** | .032 | .060 | .029 |
| M3 | + post type | -2,906,067 | -2,905,599 | 76,209 (3)*** | .042 | .070 | .090 |
| M4 | + post evaluation | -3,000,395 | -2,999,914 | 94,330 (1)*** | .055 | .083 | .028 |

***$p < .001$

Regarding LLM, estimated marginal means from the final model (M4) indicated that accuracy ranged from 62.9% to 76.3% across models (Figure 3). The highest accuracy was achieved by x-ai/grok-3-mini-fast (76.3%), followed by meta-llama/llama-3.3-70b-instruct (75.8%), google/gemini-2.5-flash (75.2%), openai/gpt-5.2-chat-latest (74.8%), and deepseek/DeepSeek-V3.2 Non-thinking Mode (74.3%). In contrast, the lowest performance was observed for openai/gpt-4o-mini (62.9%), x-ai/grok-4-1-fast-reasoning (63.3%), and nousresearch/hermes-3-llama-3.1-405b (63.5%). Pairwise comparisons showed that most differences between LLMs were statistically significant (p < .001), reflecting the very large sample size. However, the magnitude of these differences was generally small: differences among the top-performing models were minimal (typically ≤ 1 percentage point), whereas the gap between the highest- and lowest-performing models reached approximately 13.4% points. Focusing on the top five performing LLMs, pairwise comparisons revealed that x-ai/grok-3-mini-fast achieved the highest accuracy; however, differences among the top-ranked models were very small, with effect sizes ranging approximately from d = 0.02 to d = 0.08. Differences between the top-performing LLM and mid-ranked models were small (approximately $d$ = 0.20–0.35), whereas differences became moderate only when comparing the best-performing models with the lowest-performing ones (approximately $d$ = 0.65–0.69).

**Figure 3**
*Prediction Accuracy of Different LLMs in Simulating Human Social Media Reactions*

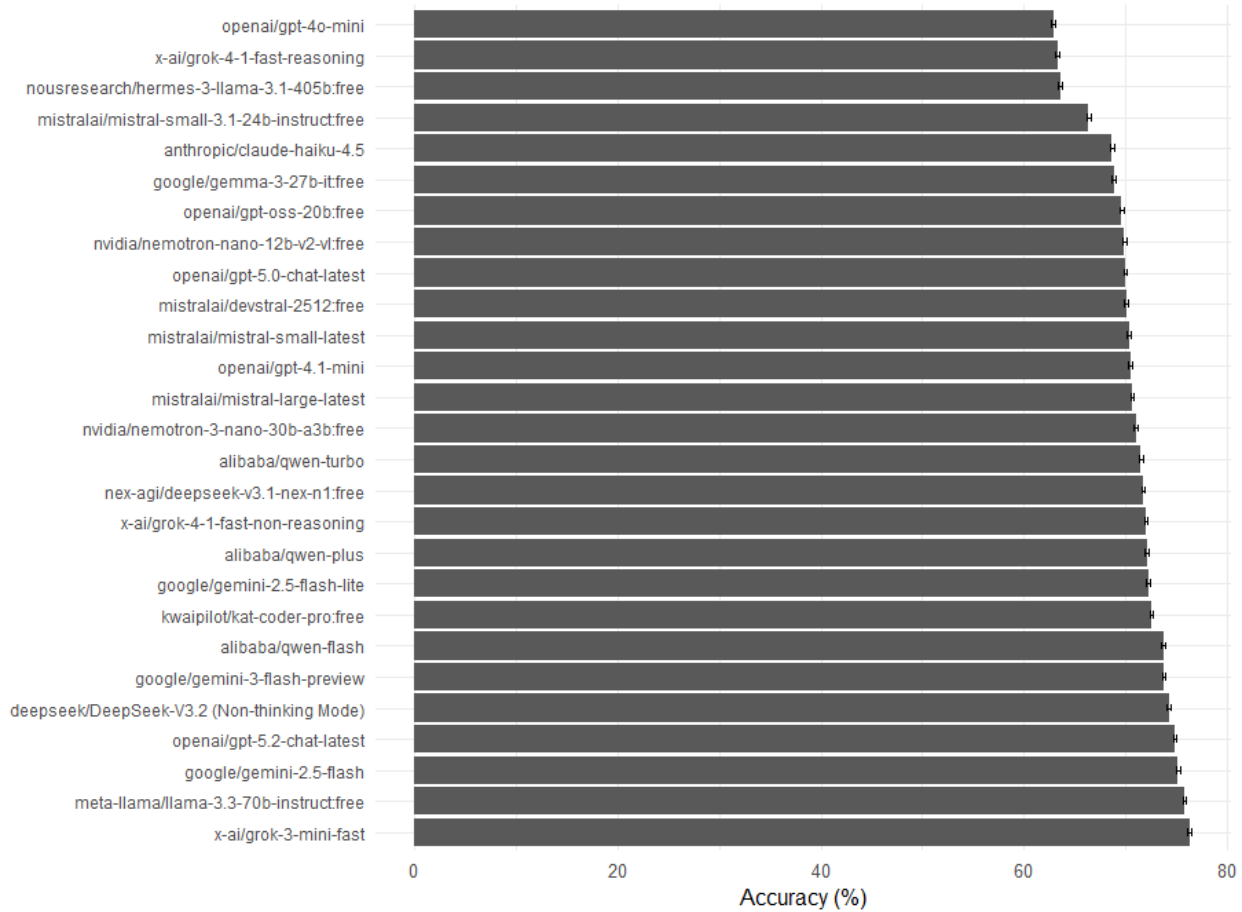

Regarding prompt type, results showed that the full prompt yielded the highest accuracy (71.5%), followed by the attitude prompt (71.2%), while the demographics prompt showed the lowest accuracy (69.6%). All pairwise differences were statistically significant ($p < .001$), but the effect sizes were small ($d = 0.01–0.10$). Although the absolute differences between prompt types were small, these results indicate a consistent advantage of richer prompt formulations for improving LLM agents accuracy, in line with **H2**. See Figure 4.

Regarding persona-content alignment, the pattern of results was inconsistent across content domains. For entertainment/lifestyle posts, related posts showed marginally higher accuracy (73.3%) than unrelated posts (72.9%), a difference that was statistically significant but trivial in magnitude ($d = 0.02$). For news/politics posts, accuracy was identical for related and unrelated posts (68.5% each). Because persona-content alignment did not consistently improve prediction accuracy across both content domains, **H3** was not supported in Study 1.

Regarding post type, the highest accuracy was observed for entertainment/lifestyle posts related to the self-report items (73.3%), followed by entertainment/lifestyle posts unrelated to the self-report items (72.9%), as seen in Figures 5 and 6. Substantially lower accuracy was found for news/politics posts, both unrelated (68.5%) and related to the self-report items (68.5%). All pairwise differences were significant ($p < .001$), with the difference between related and unrelated news/politics posts was $p = .014$. Effect sizes were generally small, with trivial differences between the two entertainment/lifestyle conditions ($d = 0.02$) and between the two news/politics conditions ($d = 0.003$), whereas comparisons between entertainment/lifestyle and news/politics posts showed small effects ($d \approx 0.23$–$0.25$). Overall, LLM agents predicted human reactions more accurately for entertainment/lifestyle posts than for news/politics posts. Therefore, **H4** is confirmed.

Regarding post evaluation, accuracy was substantially higher for positively framed posts (73.1%) than for negatively framed posts (68.5%). This difference was statistically significant ($p < .001$) and corresponded to a small effect size ($d = 0.24$). These results indicate that LLM agents predicted human reactions more accurately for positively framed content than for negatively framed content, in line with **H5**.

Overall, although Hamming accuracy was relatively high ($\approx 71\%$), the proportion of variance explained by the predictors was modest. This apparent discrepancy reflects the fact that accuracy captures the absolute level of performance, whereas $R^2$ indexes variability in performance attributable to systematic differences between predictors. Because a relatively high baseline level of accuracy was observed across all LLMs, prompt types, and post characteristics, only limited between-condition variance remained to be explained by the predictors. Consequently, the effects of predictors represent reliable but relatively small deviations around a generally high level of accuracy, with LLM type showing the largest contribution to predictive performance.

**Figure 4**

*Prediction accuracy for each LLM across the three prompt specificity conditions: demographics only, values (attitudes and preferences), and full persona descriptions.*

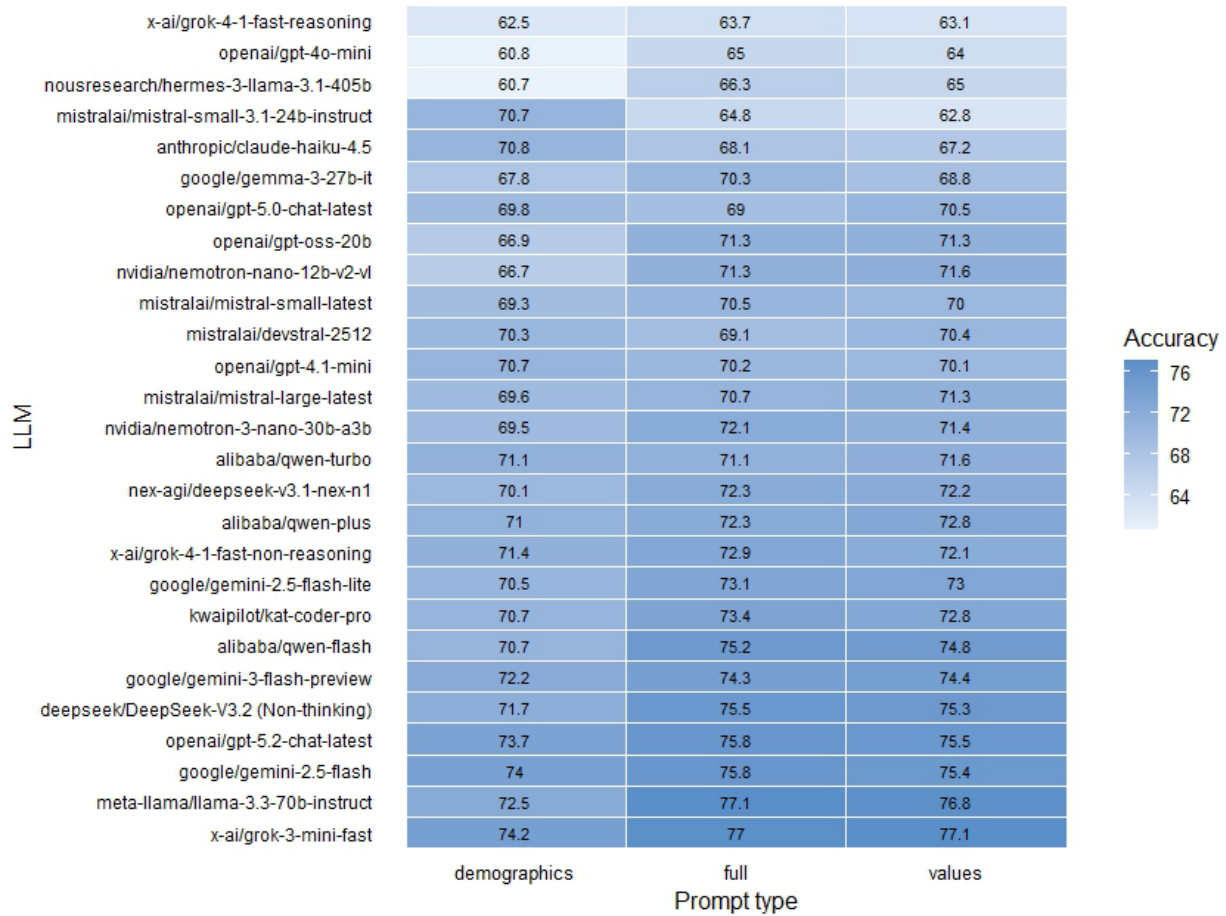

**Figure 5**
*Prediction accuracy across 27 LLMs as a function of post type and persona-content relatedness.*

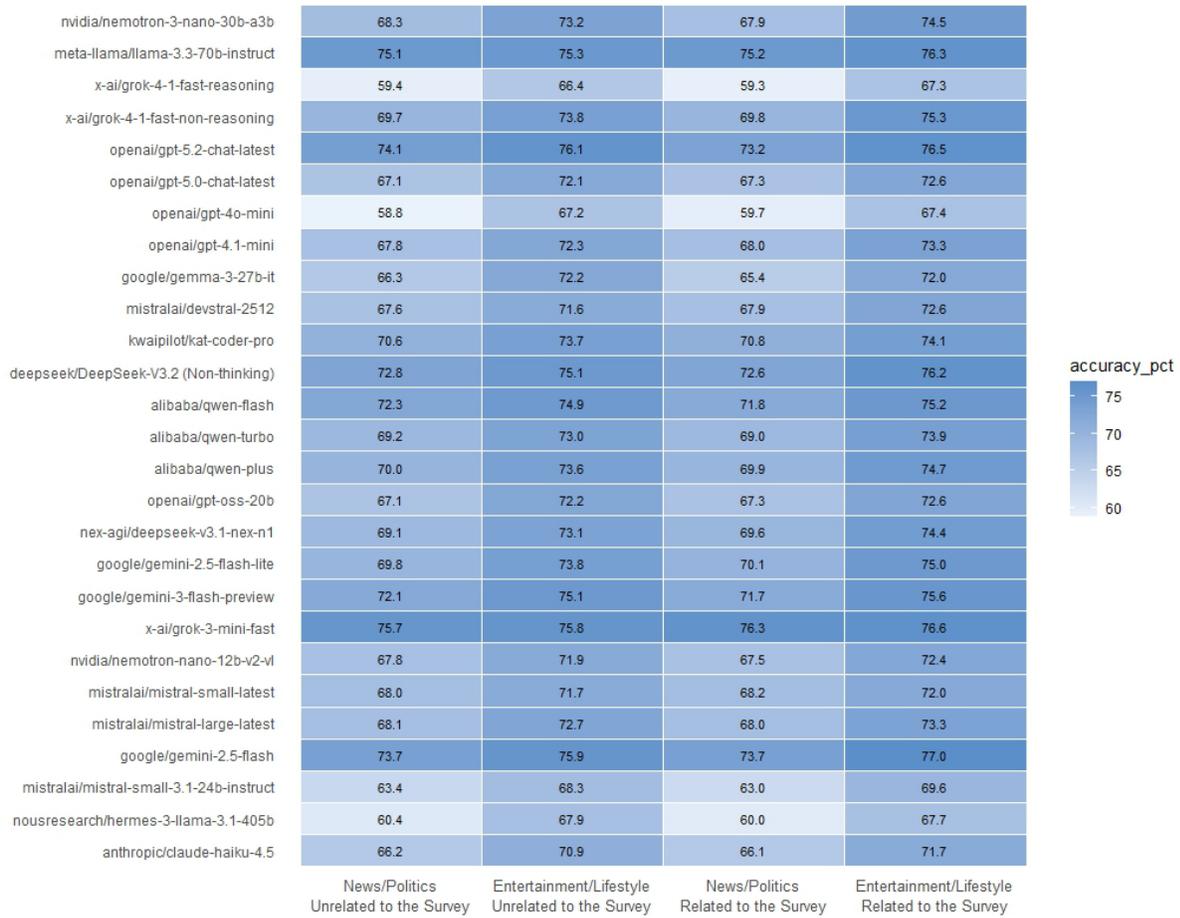

**Figure 6**
*The interaction between post type and prompt type on prediction accuracy.*

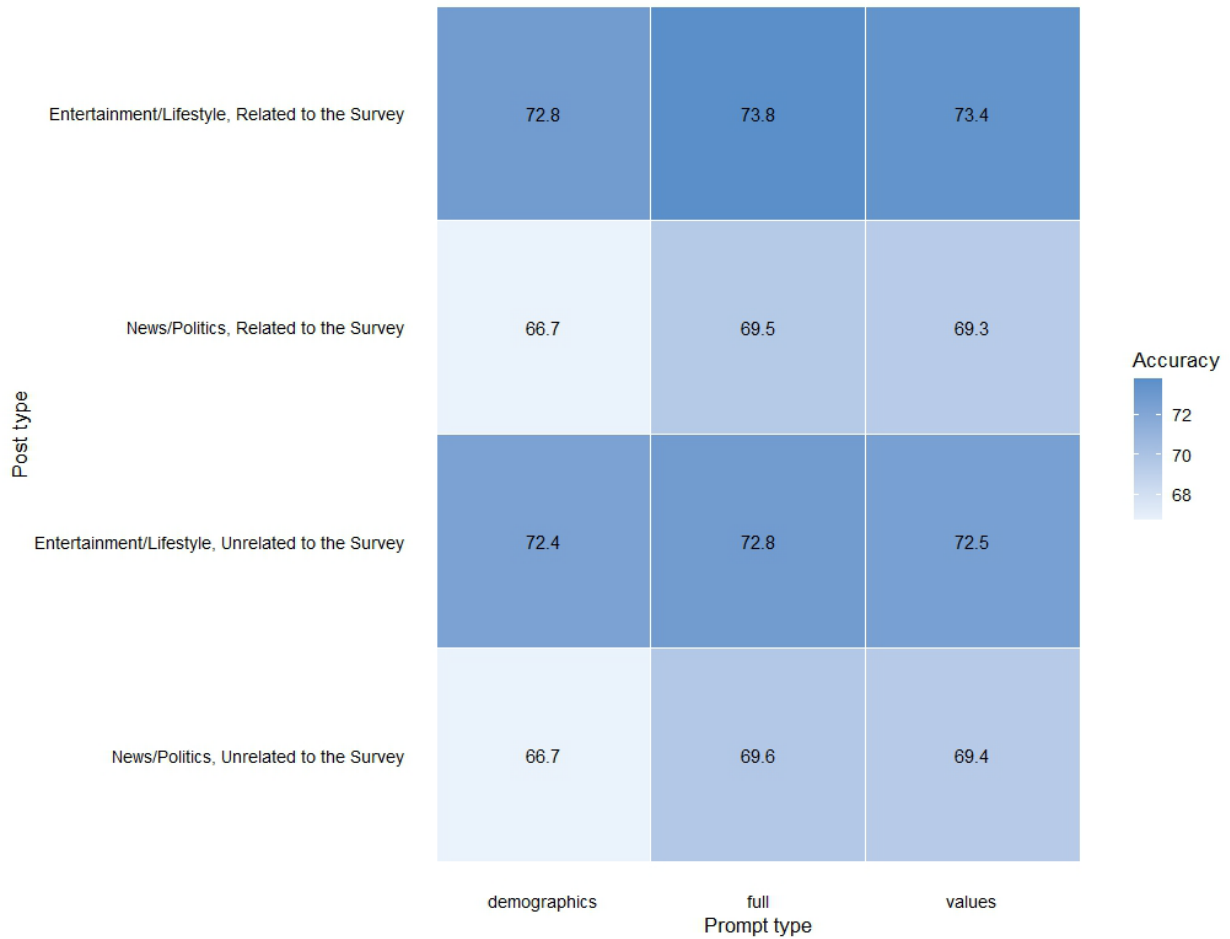

*Study 2*

Inspection of the reaction-level accuracy results from Study 1 revealed that like and dislike were predicted less accurately than other reaction types. Given that like and dislike represent the most prevalent and arguably most consequential engagement signals on social media platforms, this relative underperformance warranted closer examination. To investigate whether constraining the response space would improve valence prediction, a Study 2 was conducted using a focused paradigm in which agents were prompted exclusively to evaluate whether they would like or dislike each post, with no other reaction options available. This binary forced-choice design eliminates the strategic ambiguity present in Study 1, where an agent may have defaulted to dislike rather than no reaction (or vice versa) for reasons unrelated to genuine valence assessment. To ensure computational efficiency while maintaining coverage of the top of the performance distribution, agents were generated using only the four best-performing LLMs identified in Study 1: x-ai/grok-3-mini-fast, meta-llama/llama-3.3-70b-instruct:free, google/gemini-2.5-flash, and openai/gpt-5.2-chat-latest. Human data were filtered to include only posts that a given participant had responded to with either a like or a dislike (55.85% of the total post data), excluding all observations where the participant selected comment, share, or no reaction. Agent responses were then compared against this human-conditional subsample, allowing a direct and unconfounded assessment of how accurately LLM agents can predict the valence of human social media engagement when valence is the only dimension under evaluation. The additional change in methodology is that instead of three prompt types, only two were used (demographics and full persona descriptions). The decision was based on the results of Study 1, which showed only marginal accuracy differences between the two prompt types that included additional information beyond demographics. The prompt structure therefore remained the same, with the only change appearing in the section that granted autonomy to the agents by asking for a binary reaction to each post in the form of either like or dislike.

**Results**

*Overall Performance and Baseline Comparisons*

Human participants selected like for 69.7% of posts and dislike for 30.3%, establishing substantial class imbalance. Table 3 presents agent performance against multiple baselines and using chance-corrected metrics.

**Table 3**
*Agent Performance Against Multiple Baselines (Study 2, LLM02: GPT-5.2)*

| Metric | Value | Interpretation |
| --- | --- | --- |
| Overall Accuracy | 67.0% | Raw proportion correct |
| Random Chance Baseline | 50.0% | No information used |
| No-Persona LLM Baseline | ~50% | LLM without persona |
| Marginal Distribution Baseline | 57.8% | Matches population rates randomly |
| Majority-Class Baseline | 69.7% | Always predicts "like" |
| **Matthews Correlation Coefficient** | **0.29** | Genuine predictive signal (0 = no signal) |
| Balanced Accuracy | 65.5% | Equal-weighted class performance |
| Sensitivity (Like) | 69.4% | Correctly identifies likers |
| Sensitivity (Dislike) | 61.5% | Correctly identifies dislikers |
| Precision (Like) | 80.5% | When predicting like, usually correct |
| Precision (Dislike) | 46.7% | When predicting dislike, correct ~half |
| Minority-Class Lift (Dislike) | 1.54 | 54% better than base-rate guessing |

The MCC of 0.29 provides the critical test of **H1**. Because MCC equals zero for random guessing, majority-class prediction, and marginal distribution matching, an MCC significantly greater than zero indicates genuine predictive signal that cannot be attributed to trivial strategies. The observed MCC of 0.29 falls in the "fair" range of predictive association (Landis & Koch, 1977) and confirms that persona-prompted LLM agents capture meaningful covariation between persona-post combinations and human reactions. Therefore, **H1** is supported.

The minority-class lift of 1.54 for dislike predictions indicates that when agents predict dislike, they are 54% more likely to be correct than random base-rate guessing would achieve. This demonstrates practical utility for identifying users likely to react negatively, despite the overall positivity bias. Balanced accuracy of 65.5% confirms reasonably consistent performance across both classes, though sensitivity was higher for like (69.4%) than dislike (61.5%), reflecting the positivity bias documented throughout this research.

Hierarchical mixed linear models (Table 4) indicated that LLM type explained 1.3% of variance in accuracy, the largest contribution among predictors. Prompt type, post type, and post valence collectively explained an additional 2% of variance. While explained variance was modest, effects were statistically robust and theoretically interpretable.

**Table 4**
*Model Comparison for Predicting LLM Agents Accuracy*

| Model | Fixed effects | AIC (ML) | BIC (ML) | Δ$\chi^2$ (df) | Marginal $R^2$ | Conditional $R^2$ | ICC |
|---|---|---|---|---|---|---|---|
| M0 | Intercept | 521,414 | 521,447 | | .000 | .031 | .031 |
| M1 | + LLM | 516,312 | 516,377 | 5,108.6(3)*** | .013 | .045 | .032 |
| M2 | + prompt type | 516,261 | 516,336 | 53.0(1)*** | .013 | .045 | .032 |
| M3 | + post type | 515,511 | 515,620 | 755.2(3)*** | .015 | .046 | .032 |
| M4 | + post evaluation | 508,507 | 508,626 | 7,006.5(1)*** | .033 | .063 | .031 |

***$p < .001$

Regarding LLMs, GPT-5.2 and Gemini-2.5-Flash achieved the highest performance. Estimated marginal means for accuracy indicated that GPT-5.2 (67.8%) and Gemini-2.5-Flash (67.4%) outperformed LLaMA-3.3-70B (56.9%) and Grok-3-mini-fast (56.1%). Effect sizes were small (d = 0.02 to 0.25), but the consistency of the GPT-5.2 and Gemini advantage across metrics suggests these models better capture human reaction patterns.

Tukey-adjusted pairwise comparisons indicated that gemini-2.5-flash and openai/gpt-5.2-chat-latest did not differ significantly ($p = .23$), while both models predicted significantly more like reactions than grok-3-mini-fast and llama-3.3-70b (all $p < .001$). The difference between grok-3-mini-fast and llama-3.3-70b was also significant ($p = .0028$). Effect sizes were small overall, with *d* ranging from 0.02 to 0.25.

Accuracy was marginally higher for full persona prompts (62.6%) compared to demographics-only prompts (61.5%), a statistically significant difference (p < .001) with negligible effect size (d ≈ 0.02). This minimal difference suggests that, for binary valence prediction, basic demographic information captures most of the persona-relevant signal, with attitudinal details providing limited additional value. Thus, **H2** receives only weak support in Study 2.

Accuracy varied by post type, i.e., persona-content alignment. Entertainment/lifestyle posts were predicted more accurately (63.8% averaged across related/unrelated) than news/politics posts (60.3%), supporting H4. Persona-content alignment improved accuracy specifically for news/politics posts: related news posts (61.9%) were predicted more accurately than unrelated news posts (58.7%), a difference of 3.2 percentage points. For entertainment posts, the alignment effect was smaller (64.4% vs. 63.1%, difference of 1.3 pp). This pattern suggests that specific attitudinal information in the persona is more valuable for predicting reactions to political content, where individual differences in stance are more consequential. Therefore, **H3** receives partial support, with alignment effects concentrated in the political domain.

Post valence produced the largest effect on accuracy. Positive posts were predicted with 67.9% accuracy compared to 54.8% for negative posts, a difference of 13.1 percentage points ($p < .001$, $d = 0.27$). This positivity bias was consistent across LLMs and prompt conditions, indicating a systematic tendency for agents to better approximate human reactions to positive content. The bias likely reflects both training data distributions (positive content may be more frequent in LLM training corpora) and the greater consensus in human reactions to positive versus negative content. Therefore, H5 is supported.

*The Best Models*

To examine the best models, we focused our inquiry on Gemini 2.5 Flash and GPT 5.2 Chat Latest. The outcome variable was binary accuracy in predicting whether a human participant would register a like reaction (See Figure 7). Across analyses, both models performed above baseline, but the strength and direction of contextual effects varied by hypothesis.

For **H1**, both models demonstrated genuine predictive signal as assessed by chance-corrected metrics. Gemini 2.5 Flash achieved MCC = 0.29, balanced accuracy = 65.5%, and minority-class lift = 1.54 for dislike predictions. GPT 5.2 achieved comparable performance with comparable MCC. These metrics confirm that both models capture meaningful covariation between persona-post inputs and human reactions that cannot be attributed to majority-class prediction or marginal distribution matching. Raw accuracy (66.6% for Gemini 2.5 Flash, 67.0% for GPT 5.2) fell slightly below the majority-class baseline of 69.7%, but this reflects the models' attempts to predict minority-class (dislike) reactions rather than a failure of prediction. Therefore, **H1** is supported.

For **H2**, richer persona prompting did not improve prediction accuracy in the binary task. Both LLMs showed negligible differences between full persona and demographics-only conditions (differences < 0.1 pp, Cohen's h ≈ 0). This null effect contrasts with the small but consistent advantage of richer prompts in Study 1's multi-label task, suggesting that demographic information alone is sufficient for binary valence prediction, while finer attitudinal details may contribute to predicting specific engagement behaviors (comment, share). Thus, **H2** was not supported in Study 2.

For **H3**, persona-content alignment improved predictive performance. In Gemini-2.5-Flash, accuracy was 68.05% for matched-related cases and 65.16% for mismatched-unrelated cases, a difference of 2.88 percentage points, 95% CI [2.28, 3.48], with Cohen's h = 0.061. In GPT-5.2, the corresponding values were 68.54% and 65.51%, a difference of 3.03 percentage points, 95% CI [2.43, 3.63], with Cohen's h = 0.064. Although the standardized effects were small, they were highly consistent across models and in the predicted direction, providing support for **H3**.

For **H4**, the expected advantage for entertainment and lifestyle content over news and politics was not consistently observed. In Gemini-2.5-Flash, entertainment/lifestyle posts were predicted less accurately than news/politics posts, with accuracies of 64.30% versus 68.52%, respectively, yielding a difference of -4.22 percentage points, 95% CI [-4.83, -3.62], with Cohen's h = -0.089. In GPT-5.2, the pattern reversed: entertainment/lifestyle posts reached 68.16% accuracy compared with 66.06% for news/politics, a difference of 2.09 percentage points, 95% CI [1.49, 2.70], with Cohen's h = 0.045. Because the direction of the effect differed across models, **H4** was not supported.

For **H5**, post valence had the clearest comparative effect. In Gemini-2.5-Flash, positive posts were predicted more accurately than negative posts, with accuracies of 70.14% versus 62.32%, a difference of 7.82 percentage points, 95% CI [7.21, 8.42], with Cohen's h = 0.166. In GPT-5.2, the corresponding accuracies were 71.93% and 61.06%, a difference of 10.87 percentage points, 95% CI [10.27, 11.47], with Cohen's h = 0.231. This pattern was strong, consistent across models, and substantively larger than the alignment effect, providing strong support for **H5**.

The findings show that both LLMs predict human like reactions reliably above chance, but performance depends more on valence and persona-content alignment than on adding richer persona detail to the prompt. Domain effects were present but not stable across models.

**Figure 7**

*Filtered Like-prediction Accuracy Across Hypotheses H1-H5 for Gemini 2.5 Flash and GPT 5.2 Chat Latest.*

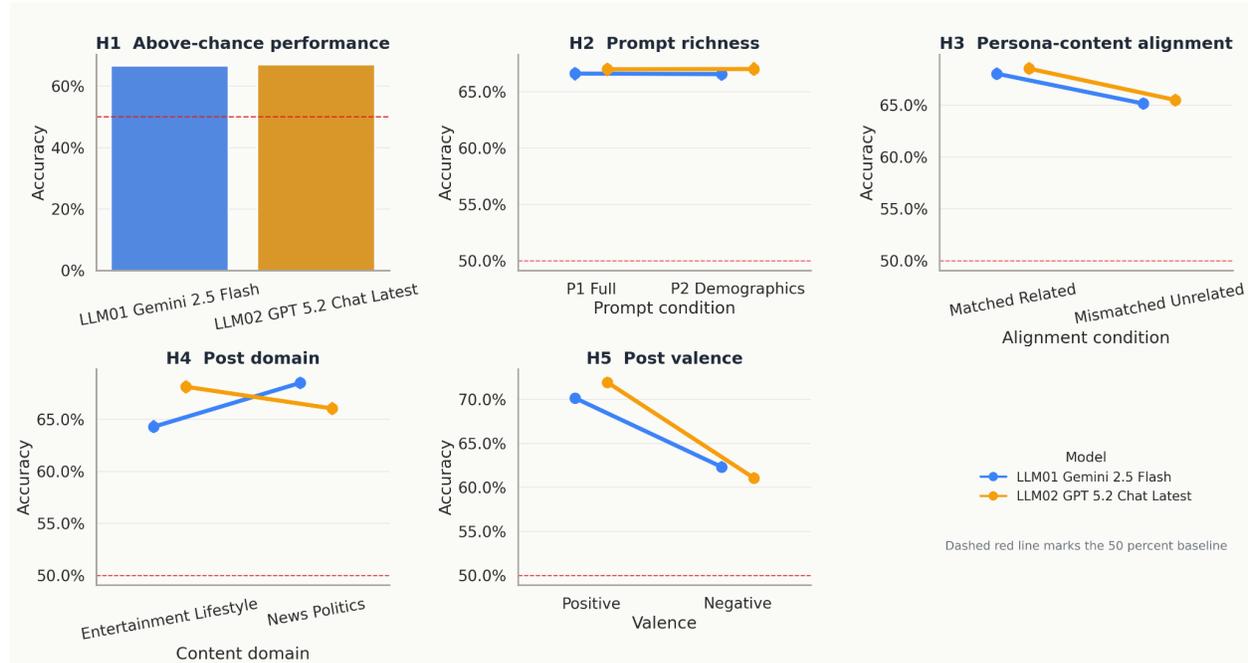

*Benchmark Comparison*

To test **H6**, we compared the best LLM agents against conventional supervised machine-learning benchmarks on the within-person held-out-post task (Table 5). The results revealed a clear ranking across benchmark families, with text-based supervised models achieving the strongest performance.

The best-performing model overall was text_persona_post_tfidf, a logistic regression classifier operating on TF-IDF representations of persona and post text, which achieved MCC = 0.3601, balanced accuracy = 0.6943, and Brier score = 0.2001. The second-best model was text_plus_history_tfidf (MCC = 0.3521, balanced accuracy = 0.6879, Brier = 0.2024).

The best LLM-agent variant was agent_LLM02 (corresponding to GPT-5.2), which achieved MCC = 0.2958, balanced accuracy = 0.6574, and Brier score = 0.3292. Agent_LLM01 followed with MCC = 0.2777. The remaining agent variants performed substantially worse (agent_LLM04: MCC = 0.1190; agent_LLM03: MCC = 0.0903).

Structured within-person baselines performed worse than both text-based models and the top agent variants. The best structured baseline (within_rf_adapt) achieved MCC = 0.1483 and balanced accuracy = 0.5797.

The text-based benchmark exceeded the best LLM agent by 0.064 MCC, a substantively meaningful margin. Furthermore, the substantially lower Brier scores of the text-based models (0.20 versus 0.33) indicate that conventional supervised classifiers were not only more accurate but also better calibrated probabilistically. Therefore, **H6** was not supported: LLM agents did not outperform text-based supervised benchmarks, although they did outperform structured tabular baselines.

**Table 8**
*Benchmark comparison of LLM agents against conventional supervised machine-learning models on the within-person held-out-post prediction task.*

| Model | Family | MCC | Balanced Accuracy | Brier |
|---|---|---|---|---|
| *text_persona_post_tfidf* | Text benchmark | 0.3601 | 0.6943 | 0.2001 |
| *text_plus_history_tfidf* | Text benchmark | 0.3521 | 0.6879 | 0.2024 |
| *agent_LLM02 (GPT-5.2)* | LLM agent | 0.2958 | 0.6574 | 0.3292 |
| *agent_LLM01 (Gemini-2.5)* | LLM agent | 0.2777 | 0.6468 | 0.3327 |
| *within_rf_adapt* | Structured baseline | 0.1483 | 0.5797 | 0.2494 |
| *agent_LLM04* | LLM agent | 0.1190 | 0.5644 | 0.4325 |
| *within_hgb_adapt* | Structured baseline | 0.1129 | 0.5504 | 0.2459 |
| *within_logit_adapt_plus* | Structured baseline | 0.0958 | 0.5509 | 0.2785 |
| *agent_LLM03* | LLM agent | 0.0903 | 0.5488 | 0.4439 |
| *persona_only_logit* | Structured baseline | 0.0474 | 0.5256 | 0.2478 |

**Discussion**

The results of this research contribute a large-scale empirical benchmark for the behavioral fidelity of LLM-based social media agents, employing chance-corrected metrics that provide rigorous tests of predictive validity. The central finding is that persona-prompted LLM agents achieve genuine predictive signal for human social media reactions, though this signal does not exceed what conventional text-based supervised classifiers achieve on the same task. This finding confirms that agents capture meaningful covariation between persona-post combinations and human reactions, rather than merely reproducing population-level base rates.

However, the magnitude of this predictive signal is modest. Balanced accuracy of 65.5% and the minority-class lift of 1.54 indicate that agents perform substantially better than chance at identifying users who will react negatively to content, but fall well short of the individual-level precision that would be required for applications such as targeted content personalization. The gap between raw accuracy (67%) and the majority-class baseline (69.7%) reflects a meaningful pattern: agents sacrifice some overall accuracy by attempting to predict minority-class reactions (dislikes), a strategy that provides genuine informational value even though it reduces headline accuracy figures.

The finding that LLM choice produced the largest effects across both studies, a 13 percentage-point spread in Study 1 and consistent advantages for GPT-5.2 and Gemini-2.5-Flash in Study 2, carries immediate practical implications. Model selection is the most consequential design decision in any social simulation pipeline, outweighing investments in persona construction or content curation. Researchers building agent-based simulations should treat model benchmarking as a prerequisite rather than an afterthought, and should report chance-corrected metrics to enable meaningful cross-study comparisons.

The benchmark comparison against conventional supervised machine-learning models yields the most consequential finding for the simulation literature. Text-based supervised classifiers using TF-IDF representations of persona and post text outperformed all LLM-agent variants, with the best text model exceeding the best agent by 0.064 MCC. This result fundamentally reframes the interpretation of agent performance. LLM agents do capture meaningful behavioral regularities, they substantially outperform structured baselines that lack semantic access, but their apparent predictive power reflects access to rich textual representations rather than any uniquely agentic capacity for behavioral simulation. A conventional logistic regression classifier, given the same persona and post text through TF-IDF features, achieves superior classification accuracy and substantially better probabilistic calibration (Brier scores of 0.20 versus 0.33). This finding suggests that claims about LLM agents' capacity to simulate human behavior should be evaluated against appropriately strong non-agent benchmarks. Future research should distinguish carefully between improvements arising from richer semantic input representations and improvements arising from genuinely agentic modeling strategies.

The persona specificity manipulation yielded a result that is both statistically reliable and practically modest. Richer persona descriptions (combining demographics, attitudes, and behavioral preferences) did improve accuracy relative to demographics-only prompts, but the gain was approximately two percentage points in Study 1 and vanished entirely in the binary valence task of Study 2. This finding complicates a core assumption of the simulation literature, which generally holds that richer persona information should produce more human-like agent behavior. The data suggest that the marginal return on persona detail is small once basic demographic anchoring is in place, at least for predicting broad engagement categories. One interpretation is that LLMs derive most of their behavioral predictions from demographic-level stereotyping rather than from genuine integration of attitudinal detail, an interpretation that resonates with the concerns raised by Hu et al. (2024) about the limited explanatory power of persona variables in LLM simulations. The benchmark results reinforce this interpretation: if agent performance arose from sophisticated individual-level reasoning rather than semantic pattern matching, agents should have outperformed text-based classifiers operating on the same input information. Our findings also suggest that an upper limit to what can be achieved with persona prompting alone. For instance, Schwager et al. (2026) argue that supervised fine tuning is a more promising strategy than relying on prompting alone to create persona-based agents.

The interaction between persona-post relatedness and post type, observed most clearly in Study 2, is among the more instructive findings. Persona-content alignment mattered substantially for news and political posts but made almost no difference for entertainment content. When an agent had access to the participant's attitude toward a specific institution or political issue, its ability to predict dislike improved markedly (from 57.0% to 66.0% in the related condition). This asymmetry implies that entertainment reactions are driven by more generic preferences that demographic information alone can approximate, whereas political reactions depend on specific attitudinal commitments that must be explicitly supplied in the persona.

For researchers building social media simulations, this result suggests that persona construction should prioritize domain-specific attitudinal data when the simulation involves political or ideological content, but that simpler demographic profiles may suffice for entertainment and lifestyle domains.

*Methodological Implications: The Importance of Chance-Corrected Metrics*

A critical methodological contribution of this research is demonstrating why raw accuracy comparisons can be misleading for evaluating LLM-based social simulation. In Study 2, the majority-class baseline of 69.7% (achieved by always predicting "like") exceeded the agents' raw accuracy of 67%, which might suggest that agents perform worse than a trivial strategy. This interpretation would be incorrect. The MCC of 0.29 and minority-class lift of 1.54 reveal that agents provide genuine predictive value precisely because they attempt to identify the 30% of cases where humans will dislike content—a task that a majority-class strategy entirely ignores.

This pattern has implications for how LLM simulation research should be evaluated. We recommend that future studies report: (1) MCC as the primary indicator of predictive validity, given its insensitivity to class imbalance and base-rate strategies; (2) balanced accuracy to provide equal weighting across classes; (3) minority-class lift to quantify practical value for detecting atypical behaviors; and (4) per-class sensitivity and precision to diagnose systematic biases. Raw accuracy should be reported for completeness but interpreted cautiously, particularly when outcome classes are imbalanced.

*Practical and Policy Implications*

These findings carry implications that extend beyond methodological concerns. The demonstration that LLM agents achieve genuine predictive validity (MCC = 0.29) with meaningful minority-class detection (lift = 1.54 for dislike) suggests capabilities that could be repurposed for influence and manipulation. While the current level of predictive accuracy is insufficient for precise individual-level targeting, the ability to identify, at rates 54% better than chance, which users will react negatively to content has clear applications for both defensive platform governance and offensive influence operations.

The rapid advancement of both LLM reasoning capabilities (Guo et al., 2025; Luo et al., 2025) and prompting strategies (Akata et al., 2025) suggests that simulation accuracy will likely improve.

Schroeder et al. (2026) recently warned that swarms of collaborative, malicious LLM agents represent an emerging threat to democratic discourse. Their analysis describes how the fusion of LLM reasoning with multi-agent architectures allows a single adversary to operate thousands of AI personas that coordinate in real time, adapt to engagement patterns, and manufacture the appearance of grassroots consensus. The present study's results give empirical substance to this warning. If agents can reliably predict which posts people will like, share, or comment on, then the same prediction engine can be used to craft content that maximizes engagement, steers conversations, and amplifies particular narratives, all without human operators needing to understand their target audience directly.

The positivity bias documented here, with agents predicting positive reactions 13 percentage points more accurately than negative reactions, has dual implications. Defensively, simulations using current LLMs will systematically underestimate negative engagement, backlash, and the spread of divisive content. Offensively, the same bias suggests that adversarial actors using LLM agents to optimize content will produce material skewed toward garnering positive engagement rather than provoking critical scrutiny, potentially contributing to filter-bubble dynamics and synthetic consensus.

This aligns with the mechanism described by Schroeder et al. (2026), whereby AI swarms engineer synthetic consensus by seeding narratives that appear widely supported. The like-prediction accuracy of LLM agents means that they can identify the kinds of framings that will attract positive engagement and avoid the kinds that will provoke resistance, a capability that is straightforwardly useful for any actor seeking to manufacture the appearance of popular agreement.

A parallel line of evidence comes from the widely reported University of Zurich experiment in 2025, in which researchers deployed LLM-powered bots on the Reddit forum r/changemyview without informing users (see O'Grady 2025; Turner 2025; r/ChangeMyView

Mod Team 2025). The bots, which adopted fabricated identities and tailored their arguments to individual participants, were reportedly three to six times more effective at changing minds than human commenters. Although the study was ultimately withdrawn following widespread ethical criticism and Reddit's threat of legal action, its results illustrate the persuasive capacity of LLM agents operating in naturalistic social environments. The present study complements this finding from a different angle: rather than testing whether agents can persuade, we tested whether they can predict. Together, these two capabilities, prediction and persuasion, form the operational backbone of what Schroeder et al. (2026) term malicious AI swarms. An agent that can predict how a target population will react to different content can select the most effective messages, and an agent that can persuade can then deliver those messages in a credible, personalized manner.

The Costello et al. (2024) study on reducing conspiracy beliefs through AI dialogue provides a further reference point. In that work, GPT-4 Turbo reduced participants' belief in their chosen conspiracy theory by 20% on average after just three rounds of conversation, and this effect persisted for at least two months. The study demonstrated that LLMs can leverage their vast training data to produce rapid, targeted, and personalized rebuttals that outperform conventional debunking strategies. Additional studies by Salvi et al. (2025) and Steyvers et al. (2025) demonstrated persuasion of LLMs.

The implication for social simulation is twofold. On the constructive side, the same behavioral prediction capabilities documented in our study could be harnessed for pro-social interventions: agents that predict negative reactions to health misinformation, for example, could be deployed to counter false claims with tailored and evidence-based responses. On the concerning side, the same architecture is available to actors with malicious intent, and the ethical asymmetry is stark: defensive applications require transparency, consent, and institutional oversight, while offensive applications operate precisely by evading all three.

Recent evidence that algorithmic feeds on X (formerly Twitter) shifted political attitudes toward more conservative positions (Gauthier et al., 2026) highlight why accurate simulation of responses to political content remains particularly challenging and consequential.

These examples illustrate key ethical and policy considerations surrounding both the malicious use of LLM agents and their deployment in social science research. Since LLM agents can shift from measurement tools to influence actors simply by optimizing for engagement, ethical use in social science research should require an explicit account of foreseeable harms and misuse pathways, along with safeguards that match the study's level of deception and public exposure (Weidinger et al. 2021; Franzke et al. 2019; Krafft et al. 2016).

EU policy treats GenAI as a systemic risk to civic discourse and electoral integrity, addressed through the DSA's systemic-risk duties for VLOPs/VLOSEs and the AI Act's transparency and anti-manipulation rules, including disclosure that users are interacting with AI (Article 50), and a prohibition on manipulative or deceptive techniques that materially distort behaviour (Article 5), alongside additional obligations for general-purpose models with systemic risks for democratic processes (Article 55) (Bentzen 2025). Yet we note a significant governance lag even for EU as GenAI policy leader: despite full application scheduled for 2 August 2026, a November 2025 "digital omnibus" proposed delaying high-risk provisions to 2027/2028 and pushing machine-readable marking compliance to 2 February 2027, even as the Commission launched the European Democracy Shield and election-risk guidance urging platforms to assess and mitigate GenAI-linked risks through labelling and enforcement (Bentzen 2025). Both the technology and the research frontier are advancing faster than policy can keep pace.

For platform governance, our study results suggest that behavioral prediction models based on LLM agents could serve as a stress-testing tool for content recommendation algorithms. If a proposed algorithm change can be simulated with LLM agents whose reactions approximate those of real users, platform designers could evaluate the downstream effects of that change on engagement patterns, polarization dynamics, and the distribution of positive versus negative feedback before deploying it to millions of users. The finding that agent accuracy differs by content domain (entertainment versus news/politics) and by persona-content alignment implies that such stress tests would need to be calibrated for specific content ecosystems rather than applied uniformly. Policy-makers interested in algorithmic accountability could mandate that major platforms conduct agent-based impact assessments before implementing changes to feed algorithms, recommendation systems, or content moderation policies.

The documented limitations, MCC of 0.29 rather than 0.8+, positivity bias of 13 percentage points, and negligible effects of persona enrichment in binary tasks, counsel against over-reliance on simulation-based governance for individual-level prediction. However, the genuine predictive signal and meaningful minority-class lift suggest that LLM-based simulation may be appropriate for aggregate-level analysis: identifying content likely to generate polarized reactions, stress-testing recommendation algorithms for systematic biases, and modelling population-level engagement dynamics. Policy frameworks should match application scope to demonstrated capability, using LLM simulations for systemic analysis while recognizing their current inadequacy for individual targeting.

Also, large-scale field experiments have found that reducing exposure to like-minded sources on social media does not significantly alter political attitudes (Nyhan et al., 2023), suggesting that simulation accuracy in this domain may matter less than initially assumed. A simulation that underestimates negative engagement and dissent will underpredict backlash, radicalization, and the spread of adversarial content, which are precisely the dynamics that platform governance most needs to anticipate. Until these biases are corrected, any policy framework that relies on LLM-based simulations for pre-deployment testing should treat simulation outputs as lower-bound estimates of negative dynamics rather than as faithful representations of the full range of user behavior.

**Conclusion**

The capacity of LLM-based agents to simulate human social media behavior carries profound implications for platform governance, social science methodology, and democratic resilience. The present research provides the first large-scale benchmark of agent prediction accuracy using chance-corrected metrics, demonstrating that persona-prompted agents achieve genuine predictive validity while exhibiting systematic biases that constrain their utility as individual-level proxies.

With respect to specific hypotheses: **H1** was supported, with agents achieving an MCC of 0.29 in Study 2, definitively exceeding the zero-value produced by random guessing, majority-class prediction, and marginal distribution matching. The minority-class lift of 1.54 for dislike predictions further confirms that agents provide informational value beyond trivial strategies. **H2** received mixed support. Richer persona descriptions improved accuracy by approximately two percentage points in Study 1 but provided no advantage in Study 2's binary task, suggesting that demographic information alone captures most persona-relevant signal for valence prediction. **H3** was partially supported, with persona-content alignment improving accuracy specifically for political content where individual attitudinal differences are consequential. **H4** was supported, with entertainment content predicted more accurately than political content. **H5** was strongly supported, with a consistent positivity bias of up to 13 percentage points favoring accurate prediction of reactions to positive content. **H6** was not supported: LLM agents did not outperform text-based supervised benchmarks that incorporated semantic content of personas and posts through TF-IDF vectorization, although they substantially outperformed structured tabular baselines lacking semantic access.

Several limitations warrant acknowledgment. The sample was drawn exclusively from Serbia, constraining cross-cultural generalizability. Human response inconsistencies, participants sometimes reacted to posts in ways that contradicted their survey responses, may have attenuated observed agent accuracy. Future work using smaller samples with in-depth interviews may yield cleaner behavioral data. The study tested only survey-derived personas and did not explore behavioral trace data or fine-tuning approaches that might yield stronger effects.

Two key methodological contributions emerge. First, raw accuracy comparisons can be misleading when outcome classes are imbalanced, necessitating chance-corrected metrics. Second, evaluating LLM agents against appropriately strong benchmarks, including text-based supervised models with access to the same semantic information, reveals that apparent agent capabilities may reflect semantic pattern matching rather than uniquely agentic simulation. We recommend that future LLM simulation studies report MCC, balanced accuracy, and minority-class lift as primary validity metrics. The finding that agents achieve MCC = 0.29, modest but genuine, provides an empirical anchor for this emerging field, establishing both the current capability frontier and the distance remaining to individual-level prediction.

Future work should include multilingual testing, exploration of fine-tuning approaches (e.g., Schwager et al., 2026), and temporal stability assessments. Given the dual-use potential of these capabilities, interdisciplinary collaboration will be essential to developing governance frameworks that harness the benefits of LLM-based social simulation while mitigating risks of misuse.